\documentclass[runningheads]{llncs}
\usepackage{graphicx}
\usepackage{tikz}
\usepackage{comment}
\usepackage{amsmath,amssymb} 
\usepackage{color}
\usepackage{cite}
\usepackage{multirow}

\usepackage{caption}
\usepackage{subcaption}
\usepackage{booktabs}
\usepackage[colorlinks, linkcolor=blue]{hyperref}
\usepackage[accsupp]{axessibility}  
\usepackage[T1]{fontenc}

\begin{document}
\pagestyle{headings}
\mainmatter
\def\ECCVSubNumber{3977}  

\title{Learn From All: Erasing Attention Consistency for Noisy Label Facial Expression Recognition} 

\titlerunning{Learn From All}
%
\author{Yuhang Zhang\orcidID{0000-0003-4161-5020} \and
Chengrui Wang\orcidID{0000-0003-0618-0797} \and
Xu Ling\orcidID{0000-0002-3495-9434}  \and
Weihong Deng\orcidID{0000-0001-5952-6996}}

\authorrunning{Y. Zhang et al.}
%
\institute{Beijing University of Posts and Telecommunications, Beijing, China\\
\email{\{zyhzyh, crwang, lingxu, whdeng\}@bupt.edu.cn}}
\maketitle

\begin{abstract}
Noisy label Facial Expression Recognition (FER) is more challenging than traditional noisy label classification tasks due to the inter-class similarity and the annotation ambiguity. Recent works mainly tackle this problem by filtering out large-loss samples. In this paper, we explore dealing with noisy labels from a new feature-learning perspective. We find that FER models remember noisy samples by focusing on a part of the features that can be considered related to the noisy labels instead of learning from the whole features that lead to the latent truth. Inspired by that, we propose a novel Erasing Attention Consistency (EAC) method to suppress the noisy samples during the training process automatically. Specifically, we first utilize the flip semantic consistency of facial images to design an imbalanced framework. We then randomly erase input images and use flip attention consistency to prevent the model from focusing on a part of the features. EAC significantly outperforms state-of-the-art noisy label FER methods and generalizes well to other tasks with a large number of classes like CIFAR100 and Tiny-ImageNet. The code is available at https://github.com/zyh-uaiaaaa/Erasing-Attention-Consistency.

\end{abstract}

\section{Introduction}

Facial Expression Recognition (FER) has wide applications in the real world, such as driver fragile detection, service robots, and human-computer interaction~\cite{she2021dive}. The most common paradigm for FER is the end-to-end supervised manner, whose performance largely relies on the massive high-quality annotated data. However, collecting large-scale datasets with fully precise annotations is usually expensive and time-consuming, sometimes even impossible. Furthermore, facial expression images have inherent inter-class similarity (all classes are human faces) and annotation ambiguity (some expression images are quite confusing), making noisy label FER more challenging than traditional noisy label classification tasks. On the other hand, it is well-known that deep neural networks have enough capacity to memorize large-scale data with even completely random labels, leading to poor performance in generalization~\cite{zhang2021understanding, arpit2017closer, jiang2018mentornet}. Therefore, robust FER with noisy labels has become an essential and challenging task in computer vision~\cite{zeng2018facial, wang2020suppressing, chen2020label, zhang2021weakly, zhang2021relative, she2021dive, gera2021noisy, fan2020learning, jiang2021boosting}. 

Mainstream noisy label FER methods can be mainly classified into two categories, sample selection and label ensembling. SCN~\cite{wang2020suppressing} and RUL~\cite{zhang2021relative} can be viewed as sample selection methods, which learn more from clean samples and then relabel the noisy samples. SCN~\cite{wang2020suppressing} uses a fully-connected layer to learn an importance weight for each sample and suppresses uncertain samples during the training phase. RUL~\cite{zhang2021relative} learns uncertainty weights through comparison between different samples. IPA2LT~\cite{she2021dive} and DMUE~\cite{she2021dive} are label ensembling methods, which provide several labels for a single sample to better mine the latent truth. IPA2LT~\cite{she2021dive} assigns each sample more than one labels with human annotations or model predictions while DMUE~\cite{she2021dive} uses a multi-branch model to better mine the latent distribution in the label space. All the aforementioned methods get good performances under noisy label FER while they still have defects. Specifically, sample selection methods are based on the small-loss assumption~\cite{zhang2021understanding, arpit2017closer}, which might confuse hard samples and noisy samples as both of them have large loss values during the training process. Sample selection methods also need the noise rate, which is non-trivial in large-scale real-world datasets. Label ensembling methods provide different views of the same sample using several networks, similar to crowdsourcing in real FER applications. However, the extra information gain they bring might be noisy. Label ensembling methods might bring great computation overhead, making them less preferable in real applications. Thus, the noisy label FER problem demands better methods that do not need to know the noise rate or train several models to perform well.

\begin{figure}[!t]
\centering
\includegraphics[width=1.0\textwidth]{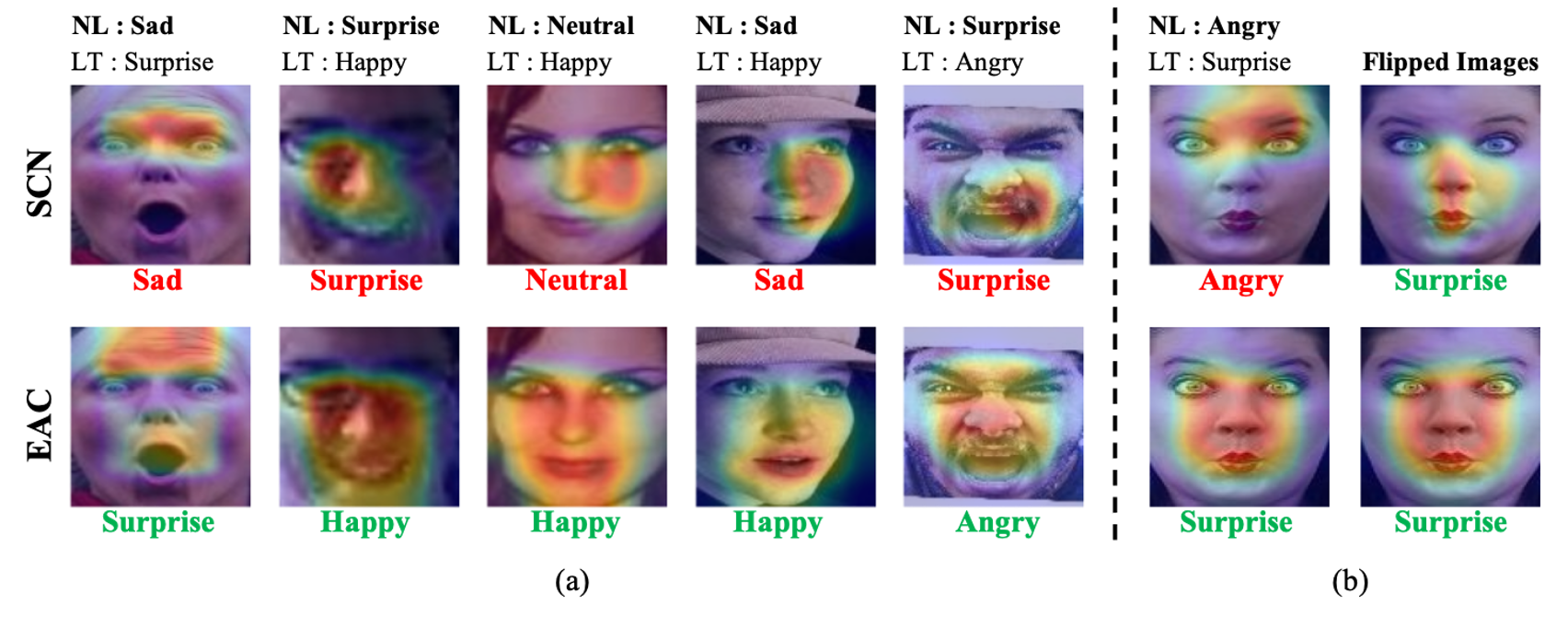}
\caption{(a) shows the attention regions of the noisy samples learned by SCN and EAC (Ours). NL represents the noisy label, LT represents the latent truth. The prediction results are shown under the images. SCN only focuses on a part of the features that can be considered related to the noisy labels to remember the noisy samples. (b) shows SCN predicts differently on the flipped image. Our EAC forces the model to focus on similar parts before and after the flip to prevent the model from remembering noisy labels.}
\label{fig:attention}
\end{figure}

In this paper, instead of following the traditional path to detect noisy samples according to their loss values and then suppress them, we view noisy label learning from a new feature-learning perspective and propose a novel framework to deal with all the aforementioned defects. We find that the FER model remembers noisy samples by focusing on a part of the features that can be considered related to the noisy labels, shown in Figure~\ref{fig:attention}. The image in the first column is labeled as sad, while its latent truth is surprise. SCN~\cite{wang2020suppressing} remembers this noisy sample by focusing on the frown feature which can be considered related to the noisy label of the sad expression. However, it neglects the open mouth feature, which is vital for the correct classification as an open mouth combined with a frown leads to the latent truth surprise instead of the noisy label sad. From the attention regions of the noisy samples, we conclude that the FER model only observes a part of the features that can be considered related to the noisy labels to remember noisy samples. It is intuitive as remembering noisy samples by focusing on a part of the features that can be considered related to the noisy labels does not contradict the other learned features from the clean samples. Inspired by this finding, we propose to deal with noisy label FER from a new feature-learning perspective. If the model can not focus on a part of the features and always learns from the whole features, then it cannot remember the noisy samples. Learning from the whole features from all training samples also means the model does not need to filter out large-loss samples like traditional methods which might confuse useful hard samples with noisy samples.

In this paper, we use Attention Consistency to implement the consistency regularization. Attention Consistency~\cite{guo2019visual} assumes that the learned attention maps should follow the same transformation as the input images to achieve better multi-label classification performance. The attention maps denote the features that the model based on to make the predictions. 

We find that the flip semantic consistency of facial expression images can help to detect noisy labels. Flip semantic consistency means the original image and its flipped counterpart should be classified into the same category. However, if we train a FER model with a noisy sample, the model might remember the noisy sample while it still predicts the latent truth on its flipped counterpart, shown as the images in the first row of Figure~\ref{fig:attention}. Inspired by that, we propose an imbalanced framework to prevent the model from remembering noisy samples. Specifically, we \emph{only} compute classification loss on the original images and compute consistency loss between the attention maps extracted from the original images and their flipped counterparts. We utilize the consistency loss to prevent the model from remembering a part of the features of the original images. Such an imbalanced framework cannot help the model totally get rid of the noisy labels as the model can still gradually overfit the attention maps of the flipped images to keep the consistency loss small, which degrades the regularization effect. We further propose Erasing Attention Consistency (EAC) to increase the performance of the imbalanced framework. Before flipping, we first randomly erase the input images during the whole training phase. During the training phase, the dynamic changing of the erased area ensures that the model can not simply remember the attention maps before and after the flip to get small consistency loss values. When the model starts to overfit the noisy original samples by focusing on a part of the features related to the noisy labels, the attention maps of the original images will deviate largely from the attention maps of their flipped counterparts, which will lead to large consistency loss values. We set the weight of the consistency loss larger enough to ensure the model first optimizes the consistency loss. Thus, to get small consistency loss values, the model will automatically quit overfitting the noisy samples.

The main contributions of our work are as follows:

\begin{itemize}
\item[1.]Instead of using traditional methods which deal with noisy labels from high-level small-loss selection, we cope with noisy labels from middle-level feature learning, which does not require the noise rate to perform well.
\item[2.]We propose a novel method named Erasing Attention Consistency (EAC) which automatically prevents the model from memorizing noisy samples.
\item[3.]We experimentally show that EAC significantly advances state-of-the-art results on multiple FER benchmarks with different levels of label noise. EAC also generalizes well to image classification tasks with a large number of classes.
\end{itemize}

\section{Related Work}
\label{sec:related}

\textbf{Noisy Label Learning} Learning with noisy labels has been well studied~\cite{patrini2017making, han2018masking, zhang2018generalized, thulasidasan2019combating, xu2019l_dmi, jiang2018mentornet, ren2018learning, arazo2019unsupervised, han2018co, malach2017decoupling, wei2020combating, xie2021partial, yi2019probabilistic, kim2019nlnl, huang2019o2u, han2019deep, li2020dividemix, ye2020purifynet, nguyen2019self, li2021learning}. Current works can be mainly categorized into two groups: modifying the primary loss function or selecting clean samples for training. 

The first type of method mainly focuses on estimating the noise transition matrix or proposing robust loss functions. Patrini \emph{et al.}~\cite{patrini2017making} estimate the transition matrix to model the relationship between noisy labels and the latent truth to prevent the model from overfitting noisy labels. Han \emph{et al.}~\cite{han2018masking} propose a human-assisted approach that conveys human cognition of invalid class transitions to make estimating transition matrix easier. Both Thulasidasan \emph{et al.}~\cite{thulasidasan2019combating} and Zhang \emph{et al.} ~\cite{zhang2018generalized} propose generalized cross-entropy loss functions to combat noisy labels. Xu \emph{et al.}~\cite{xu2019l_dmi} design a new loss function based on mutual information which is information-monotone and robust to various kinds of label noise. Although these methods have theory guarantees, they are not suitable for challenging real-world settings or handling a large number of classes. Thus, recent works usually focus on the second type of method.

The second strand of approach is based on the memorization effect that DNNs fit the underlying clean distribution before overfitting the noisy labels~\cite{arpit2017closer}. They focus on reweighting or sample selection to suppress noisy samples. Jiang \emph{et al.}~\cite{jiang2018mentornet} train a mentor net using clean samples to guide the student net by weighing the samples. Ren \emph{et al.}~\cite{ren2018learning} reweight samples according to their gradient directions. Arazo \emph{et al.}~\cite{arazo2019unsupervised} model per-sample loss by a mixture model to calculate a weight for each sample. Han \emph{et al.}~\cite{han2018co} train two models to select small loss samples for each other hoping to filter different types of error introduced by noisy labels. Malach \emph{et al.}~\cite{malach2017decoupling} improve co-teaching by updating only on instances with different predictions to keep the two models diverged. Wei \emph{et al.} ~\cite{wei2020combating} train two models together and use their agreement degree to select small-loss samples. These methods select small-loss samples to eliminate the bad influence from the noisy samples. However, the useful hard samples are likely to have large loss values and might be filtered out as noisy samples. These methods also need to know the noise rate to get better performance. Different from them, our method automatically prevents the model from memorizing the noisy samples, which do not require the noise rate or selecting clean samples.

\noindent\textbf{Facial Expression Recognition} Facial Expression Recognition (FER) aims at helping computers to understand human behavior or even interact with a human by recognizing human expression. In recent years, as the recognition accuracy is very high in the laboratory collected FER datasets, more attempts try to address the in-the-wild FER problem, which contains lots of label noise. Zeng \emph{et al.}~\cite{zeng2018facial} first consider annotation inconsistency and assign each sample with more than one label to better mine the latent truth. Wang \emph{et al.}~\cite{wang2020suppressing} propose to learn an importance weight for each sample and suppress the uncertain images by relabeling. She \emph{et al.}~\cite{she2021dive} train multi-branch models by leaving out one class for each branch in order to find the latent truth under label noise. Zhang \emph{et al.}~\cite{zhang2021relative} propose to learn the uncertainty of different facial images by comparison and then suppress the uncertain images. They can be mainly categorized into two classes, sample selection~\cite{wang2020suppressing, zhang2021relative} or label ensembling~\cite{zeng2018facial, she2021dive}. Sample selection methods select good samples and suppress noisy samples while label ensembling methods use crowdsourcing to improve performance. However, they either require the noise rate to better filter out noisy samples or bring extra computation overhead and cannot generalize well to classification tasks with a large number of classes. Our method automatically prevents the model from overfitting the noisy samples without the noise rate and generalizes well to classification tasks with a large number of classes.

\section{Proposed Method}
In this section, we illustrate the implementation details of our proposed Erasing Attention Consistency (EAC) method.
\subsection{Preliminary}
\noindent\textbf{Class Activation Mapping} Class Activation Mapping (CAM)~\cite{zhou2016cvpr} is an attention method, which allows us to visualize the predicted class scores on the given images, highlighting the discriminative parts detected by the CNN.

In the CNN trained for classification, an attention map is the weighted sum of the feature maps from the last convolutional layer with the weights from a fully connected (FC) layer. By viewing the attention maps, we can know what the model is based on to make the predictions. We denote the feature map extracted from the last convolutional layer as $ \textbf{F} \in \mathbb{R}^{C\times H\times W} $, $C$, $H$, $W$ respectively represent the number of channels, height, width of the feature map. We denote the weights of the FC layer as $ \textbf{W} \in \mathbb{R}^{L\times C} $, $L$ represents the number of classes. The attention map computes as \begin{equation} \textbf{M}_j(h, w) = \sum_{c=1}^C \textbf{W}(j,c)\textbf{F}_{c}(h, w),\label{CAM}\end{equation}
$\textbf{M}_j(h, w)$ is the attention value of location $(h, w)$ for class index $j$, which is the weighted sum of feature maps over different channels. In our method, we use CAM to compute the attention maps from the input images to show the features that the model attends to.

\noindent\textbf{Attention Consistency} Attention Consistency~\cite{guo2019visual} is first proposed for achieving better visual perceptual plausibility and better multi-label image classification by considering visual attention consistency under spatial transforms. It assumes that the learned attention maps of the model should follow the same transformation as the input images.

\subsection{Overview of Erasing Attention Consistency}
In this paper, we design an imbalanced framework to help the model get rid of the negative effect of the noisy labels. We notice that the facial images before and after the flip have the same semantic meaning of the facial expression. We only compute classification loss with the original images and compute consistency loss between the attention maps of the original images and their flipped counterparts to prevent the model from remembering the original images with noisy labels. Simply using this imbalanced framework can not help the model totally get rid of the negative effect from noisy labels as the model can gradually remember the flipped images to always get small consistency loss, which degrades the regularization effect. We further propose Erasing Attention Consistency to enhance the performance of our proposed imbalanced framework. Before flipping the original images to generate their counterparts, we first randomly erase the images according to ~\cite{zhong2020random}, which will generate different pairs of original images and their flipped counterparts during the training process. Thus, the model cannot remember the flipped images to get small consistency loss. If the model starts to remember the original images with noisy labels, the attention maps extracted from them will focus on a part of the features, which deviate largely from the flipped attention maps extracted from their flipped counterparts leading to the increase of the consistency loss. Thus, the consistency loss can prevent the model from remembering noisy samples.

\begin{figure}[!t]
\centering
\includegraphics[width=1.0\textwidth]{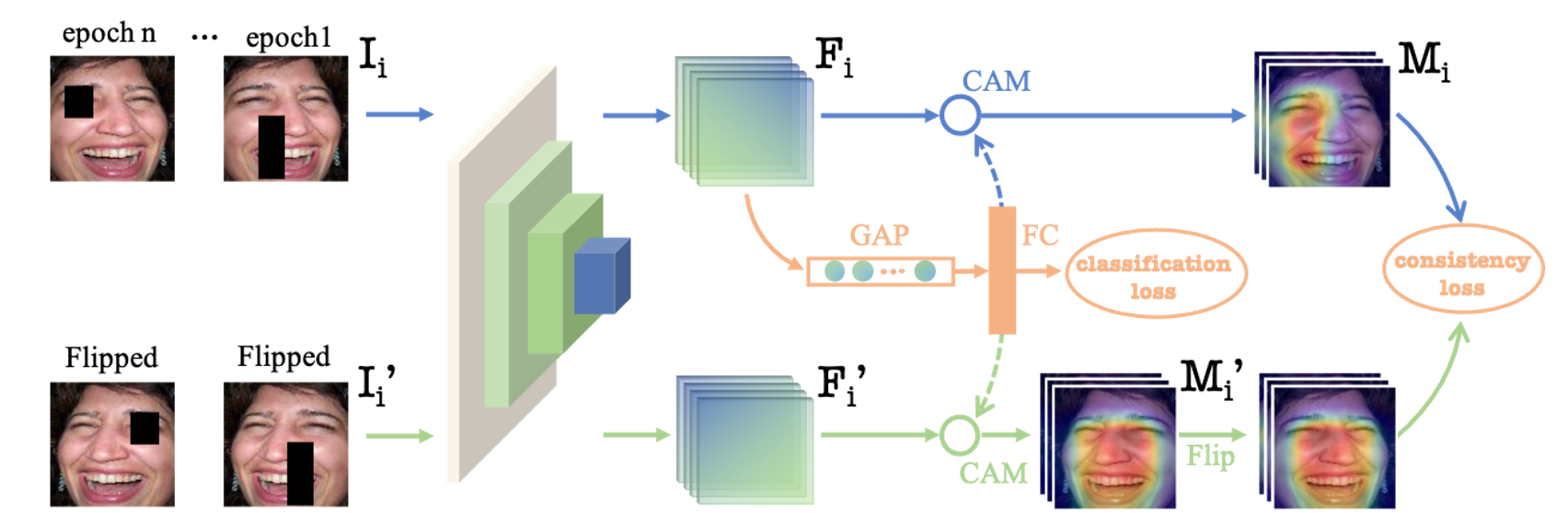}
\caption{The framework of the Erasing Attention Consistency (EAC). EAC randomly erases input images and then gets their flipped counterparts. EAC only computes the classification loss with the original images. The classification loss with the noisy labels might cause the model to overfit the noisy samples shown as $M_i$. EAC uses the consistency loss between the original images and their flipped counterparts to prevent the model from remembering noisy labels. The dotted lines mean no gradient propagation.}
\label{fig:example}
\end{figure}

\subsection{Framework of Erasing Attention Consistency}
The overall framework of our proposed EAC is shown in Figure~\ref{fig:example}. Given a batch of facial expression images, we first erase the input images according to ~\cite{zhong2020random} and get $\textbf{I}$. We then flip these images to get their flipped counterparts $\textbf{I}^{'}$. $\textbf{I}$ and $\textbf{I}^{'}$ are the input images. The feature maps are extracted from the last convolutional layer, denoted as $\textbf{F} \in \mathbb{R}^{N\times C\times H\times W}$ and $\textbf{F}^{'} \in \mathbb{R}^{N\times C\times H\times W }$. $N$, $C$, $H$, $W$ respectively represent the number of images, the number of channels, height, width of the feature maps. We \emph{only} input $\textbf{F}$ through the global average pooling (GAP) layer to get features $\textbf{f} \in \mathbb{R}^{N\times C\times 1\times 1}$. We resize features $\textbf{f}$ to $N\times C$ and put them through fully connected (FC) layer to compute classification loss according to

\begin{equation}
l_{cls} = -\frac1N \sum_{i=1}^N  (\log{\frac{e^{\textbf{W}_{\textbf{y}_i}\textbf{f}_i}}{\sum\nolimits_{j}^L e^{\textbf{W}_j\textbf{f}_i}}}), \label{classification_loss}    
\end{equation}
$\textbf{W}_{\textbf{y}_i}$ is the $\textbf{y}_i$-th weight from the FC layer with $\textbf{y}_i$ as the given label of the $i$-th image.
We compute attention maps $\textbf{M}$ and $\textbf{M}^{'}$ for $\textbf{I}$ and $\textbf{I}^{'}$ according to Eq. (\ref{CAM}). Note that the weights used to compute attention maps come from the FC layer, while the FC layer only computes classification loss with the original feature maps $\textbf{F}$. We use consistency loss to minimize the distance between the feature maps $\textbf{M}$ and $Flip(\textbf{M}^{'})$ as 
 \begin{equation} l_{c} = \frac1{NLHW} \sum_{i=1}^N \sum_{j=1}^L  ||\textbf{M}_{ij} - {Flip({\textbf{M}^{'}})_{ij}}||_2.\label{consistency loss}\end{equation}
 The total loss is computed as follows,
\begin{equation}l_{total} = l_{cls} + \lambda l_{c}.
\end{equation}
 $\lambda$ is the weight of the erasing consistency loss. The ablation study of $\lambda$ is in Section~\ref{abl}.

\section{Experiments}
In this section, we first describe 3 popular in-the-wild FER benchmarks and our implementation details. We then verify the proposed EAC on the FER datasets with different levels of label noise and study why EAC works. Visualization results of the learned features, attention maps and classification loss values are displayed to provide an intuitive understanding of EAC. We carry out an ablation study and also show the generalization ability of EAC by conducting experiments on CIFAR100~\cite{krizhevsky2009learning} and Tiny-ImageNet~\cite{russakovsky2015imagenet}. Finally, we compare EAC with other state-of-the-art FER methods.

\subsection{Datasets}

RAF-DB~\cite{li2017reliable} is annotated with basic or compound expressions by 40 trained human coders. In our experiments, images with seven basic expressions (i.e. neutral, happy, surprise, sad, angry, disgust, fear) are used including 12,271 images for training and 3,068 images for testing. 

FERPlus~\cite{barsoum2016training} is extended from FER2013~\cite{goodfellow2013challenges} with finer label annotations. It is collected by the Google search engine consisting of 28,709 training images and 3,589 test images. We use the most voting category as the annotation for a fair comparison ~\cite{barsoum2016training, wang2020suppressing, wang2020region}.

AffectNet~\cite{mollahosseini2017affectnet} is by far the largest FER dataset, which is collected from the Internet by querying expression-related keywords in three search engines containing more than one million images. There are 286,564 training images and 4,000 test images manually labeled to eight classes. 

\subsection{Implementation Details}
By default, we use ResNet-18~\cite{he2016deep} pre-trained on MS-Celeb-1M~\cite{guo2016ms} as the backbone network with the same routine as~\cite{wang2020suppressing, wang2020region, zhang2021relative, she2021dive} for fair comparisons. The facial images are aligned and cropped with three landmarks~\cite{wang2019adaptive}, resized to $224\times224$ pixels. We only use the horizontal flip and the random erasing without any other data augmentation tricks to evaluate the effectiveness of our proposed method. During training, the batch size is $256$. The initial learning rate is $0.0002$. We use Adam~\cite{kingma2014adam} optimizer with weight decay of $0.0001$ and ExponentialLR~\cite{li2019exponential} learning rate scheduler with the gamma of $0.9$ to decrease the learning rate after each epoch. The training ends at epoch $60$.

\subsection{Evaluation of EAC on Noisy FER Datasets}

We quantitatively evaluate the improvement of our proposed EAC against other state-of-the-art noisy label FER methods. We explore the robustness of EAC with three levels of label noise including the ratio of 10\%, 20\%, 30\% on RAF-DB, FERPlus, and AffectNet datasets. We follow~\cite{wang2020suppressing, zhang2021relative, she2021dive} to generate noisy labels. As the generation of label noise is random, we re-implement other state-of-the-art methods on our generated noisy datasets to make fair comparisons with them. We also consider the influence of the different backbones and backbones with or without pretraining.

Shown in Table~\ref{table:label noise}, our method outperforms other state-of-the-art FER noisy label learning methods by a large margin. For example, EAC outperforms SCN under 30\% label noise by 6.97\%, 3.24\%, 4.31\% on RAF-DB, FERPlus, AffectNet respectively. 

\begin{table}[!t]
\setlength{\tabcolsep}{3.2pt}
\begin{center}
\caption{Evaluation of EAC on noisy FER datasets. We re-implement other state-of-the-art methods and test all the methods with the same noisy datasets to make fair comparisons. Results are computed as the mean of the accuracy from the last 5 epochs}
\label{table:label noise}
\begin{tabular}{lcccc}
\hline
Method    & Noise($\%$) & RAF-DB($\%$)            & FERPlus($\%$)         & AffectNet($\%$)        \\ \hline
Baseline  & 10    &        81.01              &      83.29                &           57.24           \\
SCN (CVPR20)       & 10    &   82.15                  &      84.99              &         58.60             \\
RUL (NeurIPS21)     & 10    &  86.17 & 86.93 & 60.54 \\
EAC (Ours)     & 10    &         \textcolor{red}{88.02}             &     \textcolor{red}{87.03}               &      \textcolor{red}{61.11}                \\ \hline
Baseline  & 20    & 77.98 & 82.34 & 55.89 \\
SCN (CVPR20)    & 20    &   79.79                   &       83.35              &       57.51               \\

RUL (NeurIPS21)     & 20    &       84.32               &    85.05                  &      59.01                \\
EAC (Ours)    & 20    &      \textcolor{red}{86.05}                &      \textcolor{red}{86.07}                &     \textcolor{red}{60.29}                 \\ \hline
Baseline  & 30    &     75.50                 &       79.77               &      52.16                \\
SCN (CVPR20)     & 30    &   77.45                   &     82.20                 &       54.60               \\

RUL (NeurIPS21)     & 30    & 82.06 & 83.90 & 56.93 \\
EAC (Ours)     & 30    &       \textcolor{red}{84.42}              &      \textcolor{red}{85.44}                 &   \textcolor{red}{58.91}                  \\ \hline
\end{tabular}
\end{center}
\end{table}

Note that, unlike SCN~\cite{wang2020suppressing} and RUL~\cite{zhang2021relative}, EAC does not need to modify the labels of the training samples. Relabeling has the risk of changing right labels to wrong labels, which is less flexible than our method as EAC can automatically learn useful information from all training samples. EAC does not need to know the noise rate or tell apart hard samples and noisy samples, which fundamentally solves the defects of sample selection methods as sample selection methods require the noise rate to filter out large-loss samples, which might contain useful hard samples and useless noisy samples.

We also study EAC with different backbones. With different backbones, $\lambda$ is set to 5 under 0 and 10\% noise, 10 under 20\% and 30\% noise. As shown in Table~\ref{table:backbones}, adding EAC to MobileNet or ResNet-50 can both improve their performance. Baselines are also trained with erase and flip for a fair comparison. EAC achieves better results in all settings using ResNet-50 as backbone compared with ResNet-18 in Table~\ref{table:label noise}. The experiments of EAC using an unpretrained model as backbone are shown in the supplementary material.

\begin{table}[!t]
\setlength{\tabcolsep}{8pt}
\begin{center}
\caption{The influence of different backbones on EAC. We carry out experiments on RAF-DB. Results are computed as the mean of the accuracy from the last 5 epochs}
\label{table:backbones}
\begin{tabular}{ccccc}
\hline
Method             & 0 noise & 10\% noise & 20\% noise & 30\% noise \\ \hline
MobileNet  & 83.31\% & 77.80\%    & 70.60\%    & 62.48\%    \\
MobileNet + EAC & \textcolor[rgb]{1,0,0}{86.47\%} & \textcolor[rgb]{1,0,0}{82.63\%}    & \textcolor[rgb]{1,0,0}{81.65\%}    & \textcolor[rgb]{1,0,0}{79.82\%}    \\
\hline
ResNet-50  & 88.75\% & 83.44\%    & 79.11\%    & 71.67\%    \\
ResNet-50 + EAC    & \textcolor[rgb]{1,0,0}{90.35\%} & \textcolor[rgb]{1,0,0}{88.62\%}    & \textcolor[rgb]{1,0,0}{87.35\%}    & \textcolor[rgb]{1,0,0}{85.27\%}    \\

\hline
\end{tabular}
\end{center}
\end{table}

\subsection{Why EAC works}
We evaluate the three modules of the proposed EAC to find why EAC works well under label noise. The experiment results are shown in Table~\ref{table:modules}. Several observations are concluded as follows. Without the flip attention consistency module, the model can not use the same semantic meaning from the flipped counterparts to regularize the classification loss, which is shown in the second row. Without the erasing, the model will gradually remember the attention maps from the flipped images to get small consistency loss values, which degrades the regularization effect. Without the imbalanced framework, the noisy labels will affect the images before and after the flip together. The model can remember the noisy samples before and after the flip together, making the consistency loss useless. However, when we combine the three modules, the performance skyrockets. 

We believe it is the dynamic erasing that prevents the model from remembering the attention maps. Thus, the model needs to learn flip consistent features to minimize the consistency loss. As we only compute the classification loss with the original images (the imbalanced framework), if the model tries to remember the noisy samples, the features learned from these samples will deviate largely from their flipped counterparts, making the consistency loss large. As we set the weight of the consistency loss large enough, the model will first minimize the consistency loss. Thus, it will quit remembering the noisy samples.

\begin{table}[!b]
\setlength{\tabcolsep}{5pt}
\begin{center}
\caption{Evaluation of the three modules of EAC on RAF-DB with 30\% label noise}
\label{table:modules}
\begin{tabular}{cccc}
\hline
flip attention consistency & imbalanced framework & erasing & RAF-DB \\ \hline
\text{\sffamily x}   & \text{\sffamily x}  & \text{\sffamily x}  &     75.50   \\

\text{\sffamily x}  & \checkmark  & \checkmark   &  78.10 \\
\checkmark  & \text{\sffamily x}  & \checkmark  &  78.29 \\
\checkmark  & \checkmark  & \text{\sffamily x}  & 76.26  \\
\checkmark   &\checkmark  & \checkmark  & \textcolor{red}{84.42}  \\ \hline
\end{tabular}
\end{center}
\end{table}

\subsection{Whether flip and erase is sufficiently valid for EAC} 
We use flip because we need \emph{spatial transforms} to enable attention consistency following~\cite{guo2019visual}. Other spatial transforms like Rotate or Scale are not very effective for FER as FER test sets are mainly frontal faces with a similar scale. We utilize erasing as FER models fit noisy labels through remembering parts of the features. Erasing guides the model to focus on the whole feature as the remembered feature parts might be absent during the training. Other augments can not directly solve the part-view problem and are not very effective. We test them on noisy RAF-DB. Rotate and Scale is compared to Flip. Blur~\cite{shi2020towards} and AutoAugment~\cite{cubuk2018autoaugment} (AutoAug.) is compared to Erasing. AutoAugment searches and combines many kinds of augments together while it is still inferior to erasing.

\begin{table}[!t]
\setlength{\tabcolsep}{5pt}
\begin{center}
\caption{Comparison with other augmentation methods. The experiments are carried out on noisy RAF-DB.}
\label{table:augments}
\begin{tabular}{cccc|ccc}
\hline
Noise & Rotate & Scale & Flip  & Blur & AutoAug. & Erasing \\ \hline
10\% & 80.93\%  & 85.98\% &\textcolor[rgb]{1,0,0}{88.02\%} &   86.80\%   &  87.84\%   & \textcolor[rgb]{1,0,0}{88.02\%}   \\
20\% & 79.63\%  & 85.30\% & \textcolor[rgb]{1,0,0}{86.05\%} &  83.77\%    & 85.82\%    & \textcolor[rgb]{1,0,0}{86.05\%}   \\
30\% & 78.23\%  & 82.01\% & \textcolor[rgb]{1,0,0}{84.42\%} &   76.92\%   & 82.40\%   & \textcolor[rgb]{1,0,0}{84.42\%}   \\
\hline
\end{tabular}
\end{center}
\end{table}

\begin{figure}[!b]
\centering
\includegraphics[width=0.95\textwidth]{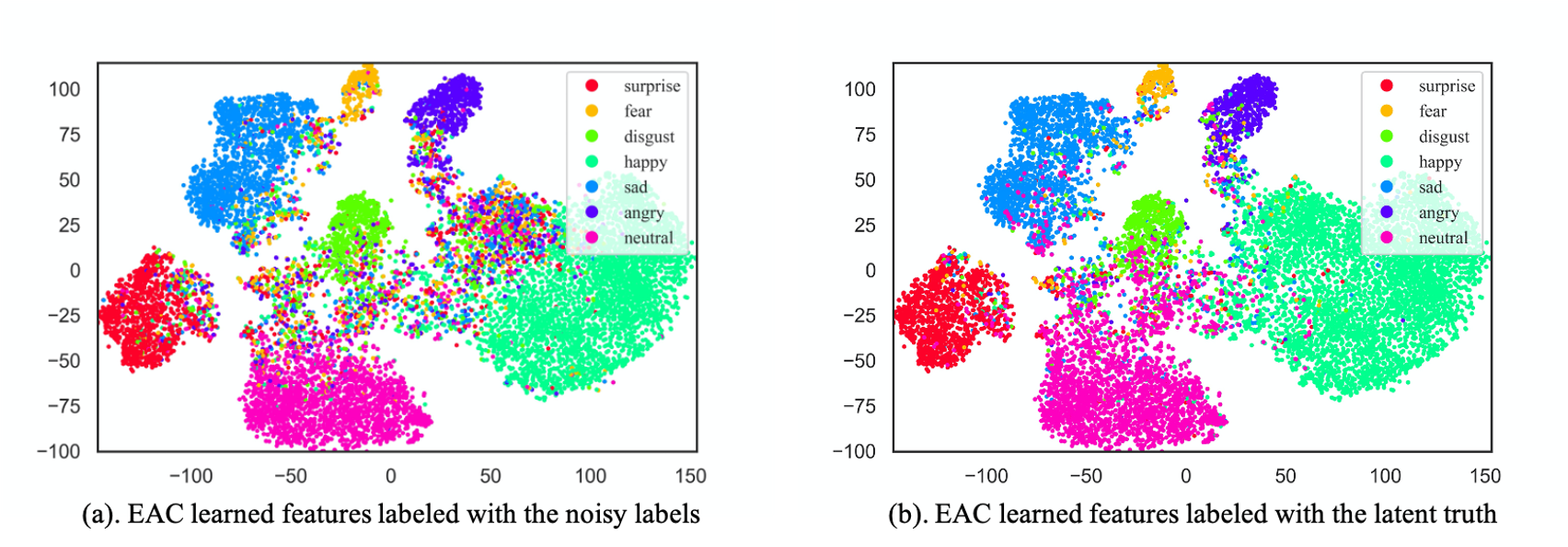}
\caption{The learned features by EAC training with noisy labels. (a) is the learned features displayed with the noisy training labels, EAC does not overfit noisy labels as different classes mixed with each other. \emph{Notice that noisy samples are pushed to the classification boundary by EAC.} (b) is the same learned features with (a), but displayed with the latent truth. Though we train EAC with noisy labels, it can still learn useful features related to the latent truth.}
\label{fig:Fvisualization}
\end{figure}

\subsection{Feature Visualization}
\label{vis}
To understand EAC intuitively, we plot the learned features of EAC trained with 30\% noisy labels on RAF-DB by t-SNE~\cite{van2008visualizing}. Figure~\ref{fig:Fvisualization} (a) is the learned features displayed with the noisy training labels. It is shown that EAC does not remember noisy labels as features with different labels are clustered together. It is shown that the features with noisy labels are close to the classification boundary which means these samples are with large classification loss values. Thus, EAC separates clean and noisy samples effectively. We also plot the same learned features in Figure~\ref{fig:Fvisualization} (b), but displayed with the latent truth. Compared with Figure~\ref{fig:Fvisualization} (a), we can draw the conclusion that EAC can automatically prevent the model from remembering noisy labels and learn useful features from both clean and noisy samples.

\begin{figure}[!t]
\centering
\includegraphics[width=1.0\textwidth]{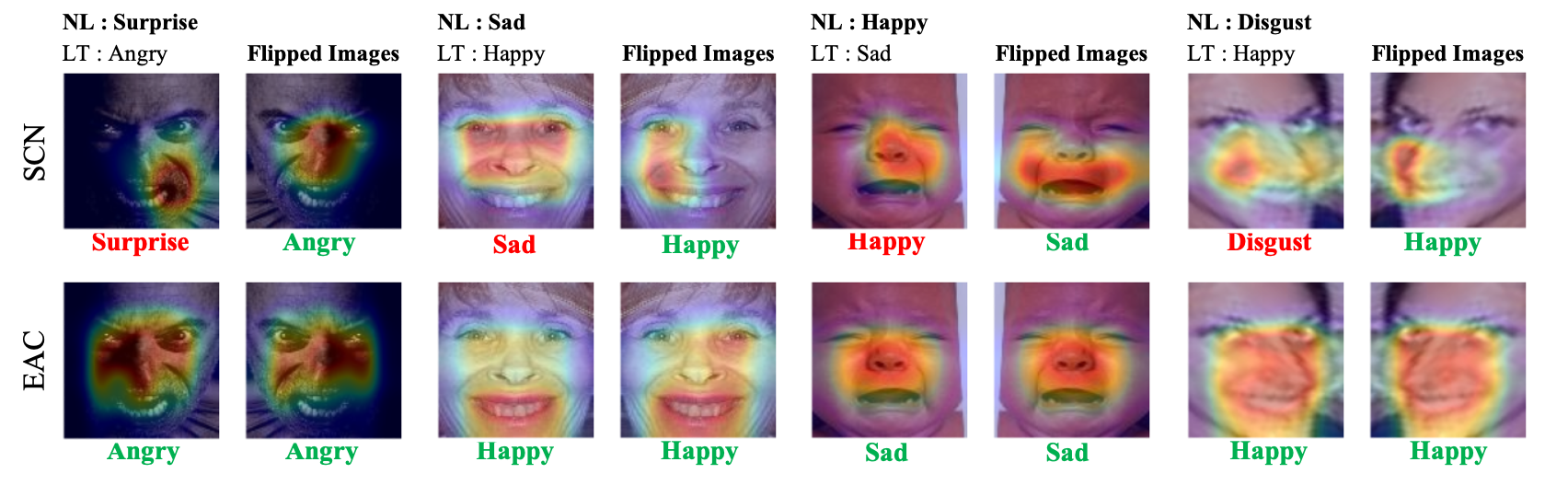}
\caption{The attention maps of SCN and EAC on the original images and their flipped counterparts.}
\label{fig:values2}
\end{figure}

We plot the attention maps on images before and after the flip in Figure~\ref{fig:values2} to show the effectiveness of EAC. We train SCN with the original images and test on their flipped counterparts. It is shown that SCN remembers the original images to the noisy labels, while it still gets correct predictions on their flipped counterparts after training. Inspired by that, EAC uses the attention maps of the flipped ones to regularize the classification loss and get correct prdictions on both the original images and their flipped counterparts. We display more results in the supplementary material.

\begin{figure}[!b]
\centering
\includegraphics[width=1.0\textwidth]{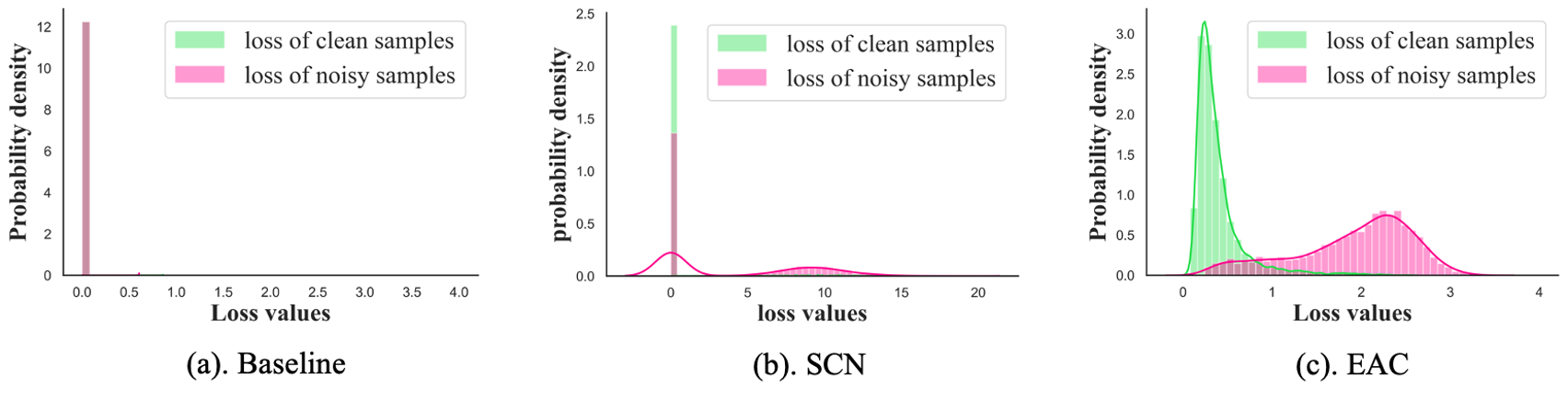}
\caption{The classification loss values of different methods after training for 60 epochs with noisy samples. The baseline remembers nearly all noisy samples. SCN avoids overfitting a part of the noisy samples, while EAC can still separate clean and noisy samples apart after training for 60 epochs.}
\label{fig:values}
\end{figure}

\subsection{Visualization of the classification loss values}
We plot the distribution of classification loss values after training for 60 epochs in Figure~\ref{fig:values} under the same setting as Section~\ref{vis}. We normalize the histogram of loss values and plot it as the probability density. The baseline method overfits nearly all the noisy samples after training for 60 epochs as the loss values of all samples are around 0. SCN learns importance weights and uses relabeling to deal with noisy samples. However, lots of the noisy samples are not correctly relabeled during the training process as there are still lots of noisy samples with loss values close to 0. Our EAC prevents the model from remembering the noisy samples during the whole training process. After training for 60 epochs, the loss values of clean and noisy samples can still be separated clearly.

\subsection{Ablation Study}
\label{abl}
We evaluate the consistency loss weight $\lambda$ from 0.1 to 10.0 with different levels of label noise. The results are shown in the supplementary material. We can choose $\lambda$ from a wide range to acquire state-of-the-art performance. The best value of $\lambda$ is 3 under 10\% and 20\% noise and 5 under 30\% noise on RAF-DB using ResNet-18 as backbone. For simplicity, we set $\lambda$ as 5 in the noisy label experiments using ResNet-18 as backbone.

\begin{table}[!b]
\begin{center}
\setlength{\tabcolsep}{1.8pt}
\caption{CIFAR100 and Tiny-ImageNet label noise training}
\label{table:generalization}

\begin{tabular}{c|llllll}
\hline
\multirow{3}{*}{Methods} & \multicolumn{3}{c|}{CIFAR100 Noise Rate}                                 & \multicolumn{3}{c}{Tiny-ImageNet Noise Rate}                             \\ \cline{2-7} 
                         & \multicolumn{3}{c|}{Top-1/Top-5 (\%)}                                         & \multicolumn{3}{c}{Top-1/Top-5 (\%)}                                          \\ \cline{2-7} 
                         & \multicolumn{1}{c}{10\%} & \multicolumn{1}{c}{20\%} & \multicolumn{1}{c}{30\%} & \multicolumn{1}{c}{10\%} & \multicolumn{1}{c}{20\%} & \multicolumn{1}{c}{30\%} \\ \hline
Baseline                 &     64.56/85.37                   &         57.33/78.93               &      49.70/72.55                  &    58.11/80.24                    &        49.56/72.43                &         41.32/64.58               \\
SCN\cite{wang2020suppressing}                    &     65.18/86.60                   &    60.38/82.11                    &         56.19/78.30               &               62.22/85.89         &        55.23/80.21                &     47.39/72.56                  \\
EAC                      &       \textcolor{red}{70.93/90.15}                 &       \textcolor{red}{66.73/87.01}                 &       \textcolor{red}{60.59/82.84}                 &    \textcolor{red}{70.22/90.23}                    &      \textcolor{red}{67.23/89.01}                  &          \textcolor{red}{63.51/87.18}              \\ \hline
\end{tabular}

\end{center}
\end{table}

\subsection{The generalization ability of EAC}

Noisy label FER methods might not be suitable for noisy label classification tasks with a large number of classes as the class number of the facial expression is very small. For example, DMUE~\cite{she2021dive} needs to train a multi-branch model whose branch number equals the class number plus 1 to mine the latent truth, which is unaffordable when the class number is very large. However, EAC can generalize well to tasks with a large number of classes.

To show the generalization ability of EAC. We carry out experiments on CIFAR100~\cite{krizhevsky2009learning} and Tiny-ImageNet~\cite{russakovsky2015imagenet}. Due to the space limitation, the implementation details are illustrated in the supplementary material. As shown in Tabel~\ref{table:generalization}, our EAC consistently improves the baseline by a large margin in both top-1 and top-5 accuracy. EAC outperforms the baseline by 6.37\% , 9.40\%, 10.89\% on CIFAR100 and 12.11\%, 17.67\%, 22.19\% on Tiny-ImageNet in top-1 accuracy with noise ratio 10\%, 20\%, 30\%. Although SCN~\cite{wang2020suppressing} also outperforms the baseline, it is clear that our EAC achieves much better results.

\subsection{Comparison with other state-of-the-art FER methods}
EAC can also help the FER model achieve state-of-the-art performance on clean datasets as EAC encourages the model to learn flip consistent features from the input images which conforms to the human visual perceptual. The results are shown in Table~\ref{table:sota}. Besides the works mentioned in Section~\ref{sec:related}, RAN~\cite{wang2020region} utilizes attention weights to aggregate a varied number of face regions to recognize facial expression robustly. DACL~\cite{farzaneh2021facial} adaptively selects a subset of signiﬁcant feature elements for enhanced discrimination. ~\cite{li2021adaptively} utilizes a knowledgeable teacher network (KTN) and a self-taught student network (STSN) to transfer knowledge. Our EAC achieves the best performance than other state-of-the-art methods on RAF-DB and AffectNet(7 classes) while slightly lower than KTN~\cite{li2021adaptively} under FERPlus. We do not compare with~\cite{xue2021transfer} as it utilizes Vision Transformer~\cite{dosovitskiy2020image} as backbone while we use ResNet-18~\cite{he2016deep}.

\begin{table}[!t]
\setlength{\tabcolsep}{6pt}
\begin{center}
\caption{Comparison with other state-of-the-art results on different FER datasets. $\dag$ denotes training with both AffectNet and RAF-DB datasets. $\ast$ denotes test with 7 classes on AffectNet.}
\label{table:sota}
\begin{tabular}{lclclc}
\hline
\multicolumn{2}{c}{RAF-DB}              & \multicolumn{2}{c}{FERPlus}             & \multicolumn{2}{c}{AffectNet} \\ \hline
Methods    & Acc. (\%)                  & Methods    & Acc. (\%)                  & Methods        & Acc. (\%)    \\ \hline
${\rm IPA2LT^{\dag}}$ \cite{zeng2018facial}       & \multicolumn{1}{c|}{86.77} &  ${\rm IPA2LT^{\dag}}$ \cite{zeng2018facial}      & \multicolumn{1}{c|}{-} & ${\rm IPA2LT^{\dag}}$ \cite{zeng2018facial}        & 57.31       \\
RAN\cite{wang2020region}        & \multicolumn{1}{c|}{86.90} &   RAN \cite{wang2020region}      & \multicolumn{1}{c|}{88.55} & RAN\cite{wang2020region}         & 59.50       \\
SCN\cite{wang2020suppressing}         & \multicolumn{1}{c|}{87.03} & SCN\cite{wang2020suppressing}          & \multicolumn{1}{c|}{88.01} &  SCN \cite{wang2020suppressing}          &  60.23       \\
DACL\cite{farzaneh2021facial}        & \multicolumn{1}{c|}{87.78} & DACL\cite{farzaneh2021facial}   & \multicolumn{1}{c|}{-} &  ${\rm DACL^{\ast}}$\cite{farzaneh2021facial}  & 65.20       \\
KTN\cite{li2021adaptively}        & \multicolumn{1}{c|}{88.07} & KTN\cite{li2021adaptively}   & \multicolumn{1}{c|}{\textcolor{red}{90.49}} &  ${\rm KTN^{\ast}}$\cite{li2021adaptively}  & 63.97      \\
DMUE\cite{she2021dive}       & \multicolumn{1}{c|}{88.76} & DMUE \cite{she2021dive}      & \multicolumn{1}{c|}{88.64} &  DMUE \cite{she2021dive}      &   62.84      \\
RUL\cite{zhang2021relative}        & \multicolumn{1}{c|}{88.98} & RUL\cite{zhang2021relative}   & \multicolumn{1}{c|}{88.75} &  RUL\cite{zhang2021relative}   & 61.43       \\

EAC (Ours) & \multicolumn{1}{c|}{\textcolor{red}{89.99}} & EAC (Ours) & \multicolumn{1}{c|}{89.64} & ${\rm EAC^{\ast} (Ours)}$     & \textcolor{red}{65.32}        \\ \hline
\end{tabular}
\end{center}
\end{table}

\section{Conclusion}

In this paper, we explore to deal with noisy label FER from a new feature-learning perspective and propose a novel and effective method named Erasing Attention Consistency (EAC). We design an imbalanced framework to utilize the erasing and flip consistency loss to prevent the model from remembering noisy labels. EAC does not require the noise rate or label ensembling. Extensive experiments verify that EAC outperforms other state-of-the-art noisy label FER methods on clean and noisy datasets. Furthermore, EAC generalizes well to noisy label classification tasks with a large number of classes.


~\\
\textbf{Acknowledgments:} This work was supported in part by the National Natural Science Foundation of China under Grant 62192784 and Grant 61871052.


%
%
\bibliographystyle{splncs04}
\bibliography{eccv2022submission}
\end{document}